\documentclass[10pt]{article}
\usepackage{setspace}
\usepackage[margin=1.1in]{geometry}

\usepackage[font=footnotesize]{caption}

\usepackage{amsgen}
\usepackage{amsfonts}
\usepackage{amssymb}
\usepackage{amsbsy}

\usepackage{graphicx}
\usepackage[breaklinks=true,colorlinks=true,linkcolor=blue,urlcolor=blue,citecolor=blue]{hyperref}
\usepackage[ruled,vlined]{algorithm2e}
\usepackage{algorithmic}
\usepackage{color}
\usepackage{multirow}

\usepackage{amsmath,amssymb}

\usepackage{caption}
\usepackage{subcaption}
\title{Deformation Driven Seq2Seq Longitudinal Tumor and Organs-at-Risk Prediction for Radiotherapy}
\author{Donghoon Lee, Sadegh R Alam, Jue Jiang, Pengpeng Zhang, \\Saad Nadeem* and Yu-Chi Hu*{\footnote{{\hspace{-4mm}*Co-senior authors.}}}}
\date{}

\usepackage{fancyhdr}
\begin{document}
\maketitle

{%\footnotesize
%\noindent
Department of Medical Physics, Memorial Sloan Kettering Cancer Center, New York, NY, USA
}\\

\begin{abstract}
\noindent \textbf{Purpose:} Radiotherapy presents unique challenges and clinical requirements for longitudinal tumor and organ-at-risk (OAR) prediction during treatment. The challenges include tumor inflammation/edema and radiation-induced changes in organ geometry, whereas the clinical requirements demand flexibility in input/output sequence timepoints to update the predictions on rolling basis and the grounding of all predictions in relationship to the pre-treatment imaging information for response and toxicity assessment in adaptive radiotherapy.

\noindent \textbf{Methods:} To deal with the aforementioned challenges and to comply with the clinical requirements, we present a novel 3D sequence-to-sequence model based on Convolution Long Short Term Memory (ConvLSTM) that makes use of series of deformation vector fields (DVF) between individual timepoints and reference pre-treatment/planning CTs to predict future anatomical deformations and changes in gross tumor volume as well as critical OARs. High-quality DVF training data is created by employing hyper-parameter optimization on the subset of the training data with DICE coefficient and mutual information metric. We validated our model on two radiotherapy datasets: a publicly available head-and-neck dataset (28 patients with manually contoured pre-, mid-, and post-treatment CTs), and an internal non-small cell lung cancer dataset (63 patients with manually contoured planning CT and 6 weekly CBCTs).

\noindent \textbf{Results:} The use of DVF representation and skip connections overcomes the blurring issue of ConvLSTM prediction with the traditional image representation. The mean and standard deviation of DICE for predictions of lung GTV at week 4, 5, and 6 were 0.83$\pm$0.09, 0.82$\pm$0.08, and 0.81$\pm$0.10, respectively, and for post-treatment ipsilateral and contralateral parotids, were 0.81$\pm$0.06 and 0.85$\pm$0.02. 

\noindent \textbf{Conclusion:} We presented a novel DVF based Seq2Seq model for medical images, leveraging the complete 3D imaging information of a relatively large longitudinal clinical dataset, to carry out longitudinal GTV/OAR predictions for anatomical changes in HN and lung radiotherapy patients, which has potential to improve RT outcomes.
\end{abstract}
%\vspace{2pc}

\section{Introduction and Purpose}
The recent development of imaging techniques in modern radiation therapy (RT) has enabled an improved understanding of the geometric variation of patient anatomy during the treatment course \cite{hugo2017longitudinal}. The innovative methods to manage this longitudinal variation fall under the umbrella of Adaptive Radiotherapy (ART)
\cite{hugo2017longitudinal,brock2017use,jaffray2005emergent}. ART is a conceptually attractive approach to correct for daily tumor and normal tissue variation. For example, ART for head and neck cancer is an emerging tool to counter morphologic changes in patient and tumor anatomy during a course of RT by creating new radiation plans mid-treatment \cite{schwartz2013adaptive,jaffray2012image,yan1998use,mencarelli2014deformable}. The most common anatomical variation is excessive weight loss and radiation-induced side effects \cite{mencarelli2014deformable,munshi2003weight,lee2008assessment}. These longitudinal anatomical changes have been shown to reduce doses to target volumes while increasing the dose to the organs-at-risk (OARs) \cite{wang2019toward,brouwer2015identifying}. 

According to a retrospective study, the time course of treatment setup variation can be characterized early on during the treatment process \cite{jaffray2012image,yan1997adaptive}. Therefore, ART may potentially benefit from patient's anatomical deformation predictions of the later timepoints since these could directly feed into the replanning process and improve the therapeutic outcome \cite{nadeem2020ldeform,lee2020predictive,wang2019toward}. In this paper, we present a novel 3D deep learning sequence-to-sequence model (Seq2Seq) using ConvLSTM to predict patient anatomy deformation (with reference to the planning/pre-treatment image) given any number of input/output sequence timepoints.

The Seq2Seq model is developed to address the history-dependent response prediction problem \cite{wang2020general} and has been used in various applications such as natural language process, image captioning, weather forecasting, and video frame prediction \cite{wang2020general,liu2017table,liu2019sequence,kim2017deeprain,kim2019deep,mukherjee2019predicting}. In the Seq2Seq model, Long Short Term Memory (LSTM) or Gated Recurrent Unit (GRU), which are representative recurrent neural network (RNN) models, have been used \cite{hochreiter1997long,cho2014learning}. LSTM and GRU are designed for the next time-step status prediction in a temporal sequence and can be naturally extended to predict the later frames from previous ones in a sequential image dataset \cite{zhang2019spatio,shi2015convolutional}. However, video or image prediction by LSTM or GRU has been combined with Fully Convolutional Networks (FCNs) in order to learn spatial and temporal filters. 3D feature maps of shape $[C,H,W]$ are flattened into 1D vectors of size $C\times W \times H$, where $C$ represents the channels, $H$ the height, and $W$ the width of the feature maps. A disadvantage of this approach is that the data that flows through the LSTM is 1D, and as such we may lose spatial information. One approach that mitigates this issue and is suitable for dealing with the sequence of images is the ConvLSTM \cite{zhang2019spatio}. It is a recurrent model, just like the LSTM, but internal matrix multiplications are exchanged with convolution operations. As a result, the data that flows through the ConvLSTM cells preserves its input dimensionality instead of being flattened to 1D feature vectors. Because of the spatio-temporal trainable advantage of ConvLSTM, it has been applied to medical image sequential datasets \cite{shi2015convolutional}. Zhang et al. \cite{zhang2019spatio} studied tumor growth prediction using ConvLSTM with cropped $32\times32$ patches of tumor CT, tumor contour, and intercellular volume fraction images. Even though the results achieved 0.80 dice overlap, their model suffered from the known blurring and limited long-term dependency ConvLSTM issues \cite{lange2020attention,tang2019multiple}. Moreover, the study focused on predicting a single timepoint from two earlier timepoints. GAN-based approaches such as BeyondMSE \cite{mathieu2015deep} have also been proposed for deep learning future video frame prediction but as was shown in \cite{zhang2019spatio}, it achieved lower prediction performance than ConvLSTM given the lack of explicit temporal dynamics modeling in BeyondMSE. Zhang et al. \cite{zhang2019spatio} also noted that the GAN-based prediction methods have a higher risk of overfitting and the network architectures can be over-complicated for the relatively small-sized medical datasets.

\noindent
In this paper, we make the following contributions:
\begin{enumerate}
    \item We present a novel 3D deep learning sequence-to-sequence model using ConvLSTM for longitudinal tumor and OAR prediction during radiotherapy given any number of input/output sequence timepoints (e.g. predicting weeks 4--6 given weeks 1--3 sequence input or predicting weeks 3--6 given weeks 1--2 sequence input) and the whole imaging field-of-view (rather than cropped patches).
    \item Deformation vector field (DVF) representation is used to overcome the ConvLSTM blurring issue with the image representation and to keep the predictions grounded with respect to the reference planning/pre-treatment images (radiotherapy clinical requirement). Skip connections are used to resolve the limited long-term dependencies ConvLSTM issue between sequence elements.
    \item A thorough validation is done on two radiotherapy datasets: (1) publicly available head-and-neck (HN) 28 patient dataset with three pre-, mid-, and post-treatment CT timepoints, and (2) internal non-small cell lung cancer (NSCLC) 63 patient dataset with planning CT and 6 weekly cone-beam CT (CBCT) images. All timepoints in both datasets were contoured for tumor and OAR by expert radiation oncologists, providing data for evaluation and for creating high-quality training DVFs via hyper-parameter optimization. For OAR, we focus on longitudinal prediction for esophagus in lung (due to the debilitating radiation-induced esophagitis which develops in 50\% of the NSCLC radiotherapy patients) and for parotid glands in HN (due to the prevalent radiation-induced xerostomia). The use of 6 weeks' sequence 3D CT images makes our work the first to study such a long sequence in the longitudinal prediction context with a relatively large dataset. 
\end{enumerate}

In the following sections, we will describe the two HN and lung datasets, the DVF creation process via deformable image registration (DIR) hyper-parameter optimization, the Seq2Seq deep learning prediction model and finally conclude with results and discussion.

\section{Materials \& Methods}
\label{sec:Materials & Methods}
\subsection{Longitudinal dataset}
We used two longitudinal radiotherapy datasets: a publicly available head-and-neck squamous cell carcinoma (HNSCC) and an internal non-small cell lung cancer datasets. The HNSCC dataset, which is available via The cancer image archive (TCIA), contains three-dimensional high-resolution 3D fan-beam CT scans collected pre-, mid-, and post-treatment for 28 patients who underwent RT treatment (to a total dose of 58--70 Gy using daily 2--2.20 Gy fraction for 30--35 fractions) \cite{bejarano2019longitudinal}; there were total 31 patients but 3 were missing parotid gland contours. The dataset also contains radiation oncologist drawn contours of anatomical structures and dose maps. The pre-treatment CT scans were acquired a median of 13 days before treatment, the mid-treatment acquired at fraction 17, and the post-treatment at fraction 30 \cite{bejarano2019longitudinal}. 
The lung dataset is an internal dataset and contains three dimensional CT and weekly CBCT scans collected during RT of 63 patients with NSCLC. Patients were treated via intensity-modulated RT in 2--3 Gy daily fractions in a five days/week fraction (total dose 50—-73 Gy) \cite{alam2020quantification}.  High-resolution planning CT (pCT) was used to define the Gross tumor volume (GTV) and planning the treatment. All patients had weekly CBCT to monitor positional uncertainties and tumor changes. In addition to imaging data, we have the contours of GTV and Esophagus at planning CT and each weekly CBCT. The contours were generated by a radiation oncologist. We split each dataset into training/validation/testing with 21/2/5 for HN and 50/3/10 for lung.
 
\subsection{Deformable image registration and hyper-parameter optimization}

The purpose of our study is to predict future sequence DVF from earlier DVF sequence so we need accurate DVF data from deformable image registration (DIR) to train our model. First, we roughly aligned the global structures by matching the center of the tumor and rigidly registering mid- and post-treatment CTs to pre-treatment CT for HNSCC dataset and weekly CBCTs to pCT for the lung dataset. Then we resampled the datasets with an isotropic voxel size, $1\times1\times1$ $mm^3$ for HNSCC and $2\times2\times2$ $mm^3$ for lung dataset, to have the same physical deformation intensity in the 3-axis direction. The HNSCC volumes were then cropped to include only the HN information, and the lung pCT volumes were cropped to match the CBCT field of view which is inherently different. Using these final images and the corresponding OAR/GTV contours, optimal hyper-parameters for B-spline regularized diffeomorphic registration \cite{avants2011reproducible} were computed using hyper-parameter optimization \cite{le2016quantifying,le2017sparse} with dice output; planning and pre-treatment CTs were used as references/targets and the weekly CBCTs/mid-/post-treatment CTs as sources for DIR, and the final transformation in the diffeomorphic registration is obtained using a Symmetrized Large Deformation Diffeomorphic Metric Mapping (LDDMM) algorithm \cite{avants2011reproducible} 
{available in Advanced Normalization Tools (ANTs.) The transformation in LDDMM maps the corresponding points between two images by finding a geodesic solution. The integrated B-spline regularization models the DVF as a B-spline object to capture large deformation. This gives free-form elasticity to the converging/diverging vectors that represent a morphological shrinkage/expansion\cite{avants2011reproducible}.}

For deciding the optimal DIR hyper-parameters, we focused on grid size and gradient step which greatly influence the final results. We investigated different combinations of grid size (16,32,64,128) and gradient step size (0.01,0.1,0.3,0.5). The grid size number denotes grid size at the lowest resolution stage in three-step multi-resolution DIR and the grid size is reduced by multiples of two as the resolution increases. The mutual information was used as a metric and the iteration time for DIR was 100, 70, and 40 for each resolution step. The dice between the target and the warped contours (GTV/parotid for HN and GTV/esophagus for lung) was used as the final output variable. The hyper-parameter optimization (solved using grid search) is formulated as follows:
%$$ x^*, y^* = arg \max_{x,y} f(x,y) $$

$$ \theta^*_{DIR} = arg \max_{\theta_{DIR}} f_{Dice}(x|\theta_{DIR}) $$

where $f_{Dice}$ is the objective function that is a combined GTV$+$OAR dice in our context, $\theta_{DIR}$ is the set of hyper-parameters in DIR (grid size and gradient step size in this study.) The purpose of hyper-parameter tuning is to retrieve optimal values that result in best DIR performance across datasets.

\begin{figure}[h!]
\begin{center}
\footnotesize
\setlength{\tabcolsep}{3pt}
\begin{tabular}{c|c|c|c}
\includegraphics[width=0.22\textwidth]{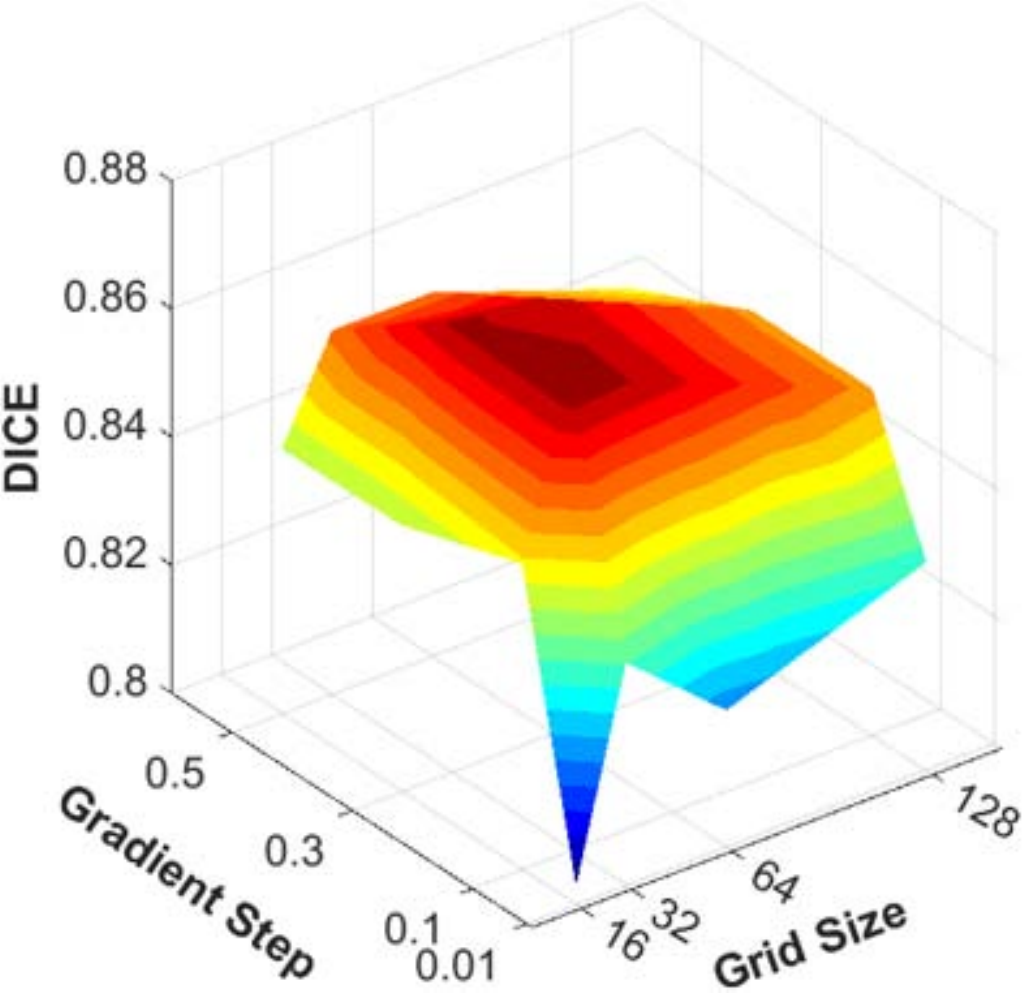}&
\includegraphics[width=0.22\textwidth]{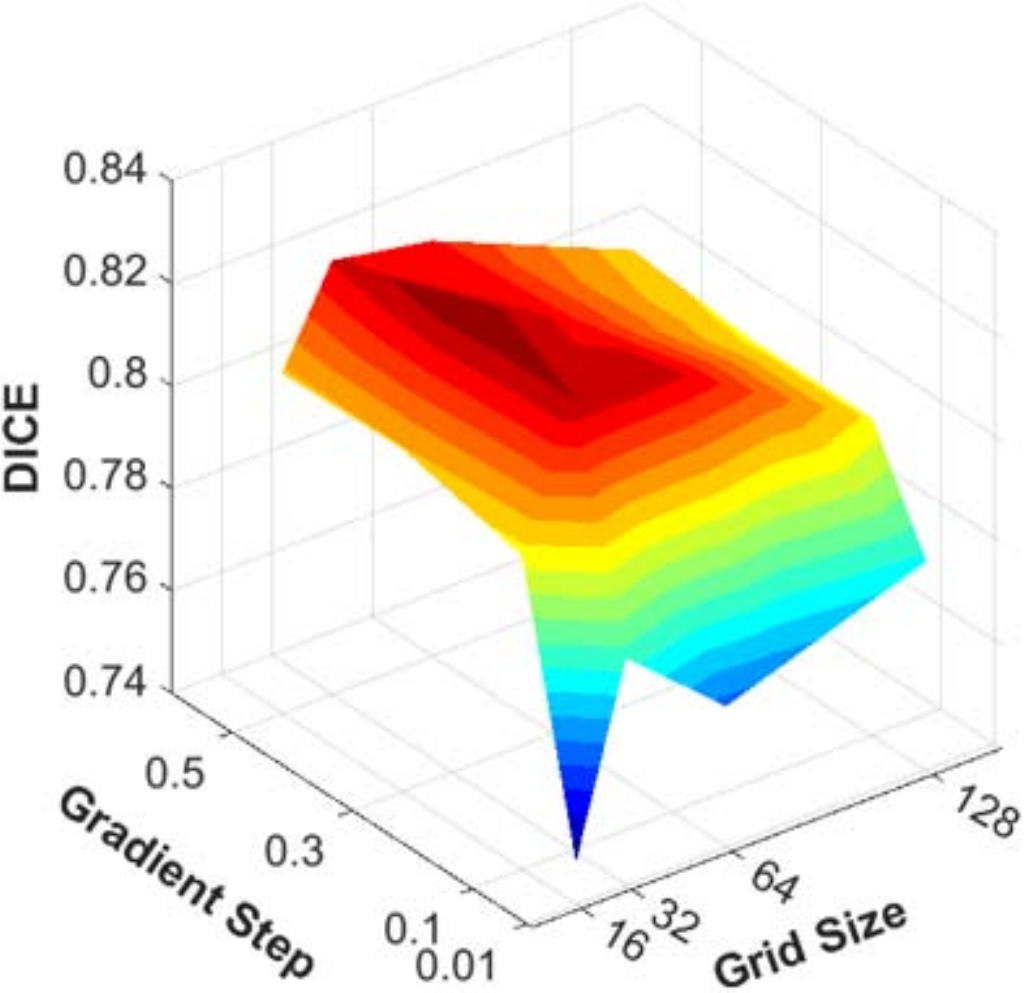}&
\includegraphics[width=0.22\textwidth]{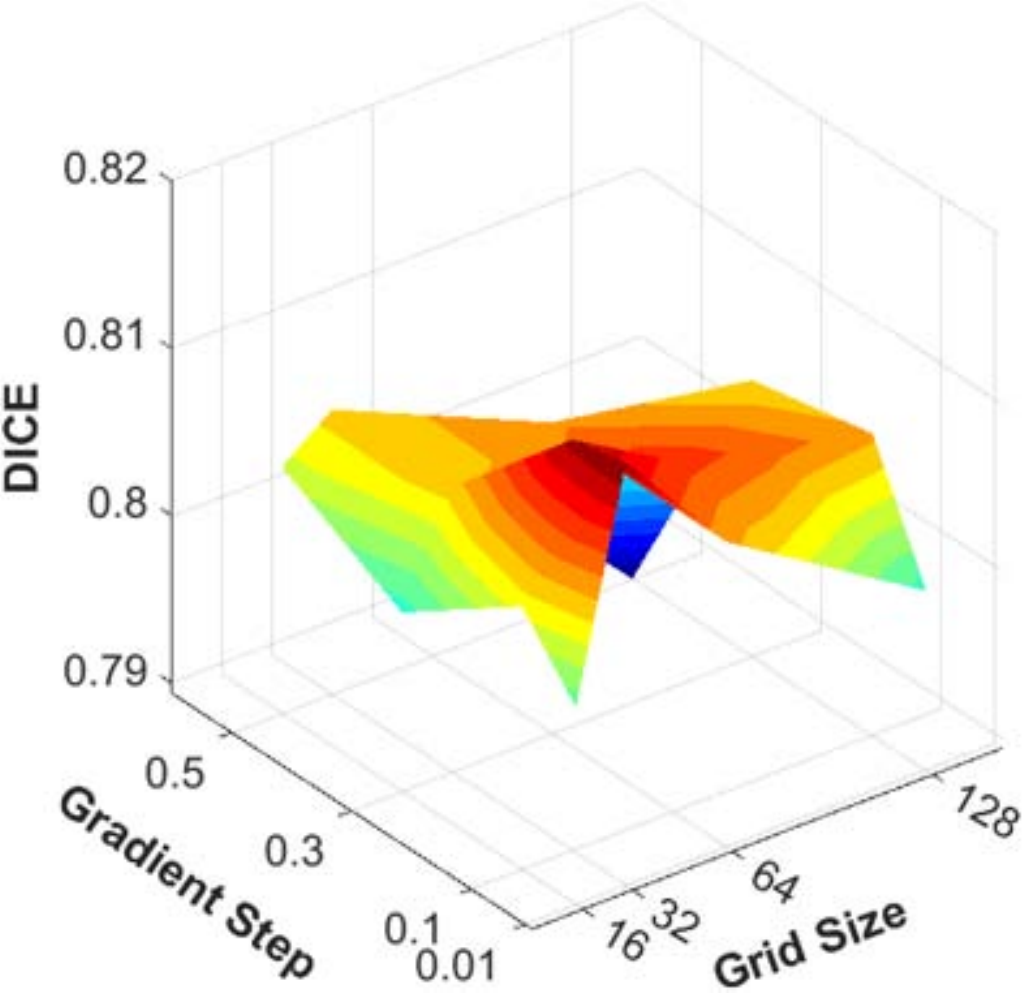}&
\includegraphics[width=0.22\textwidth]{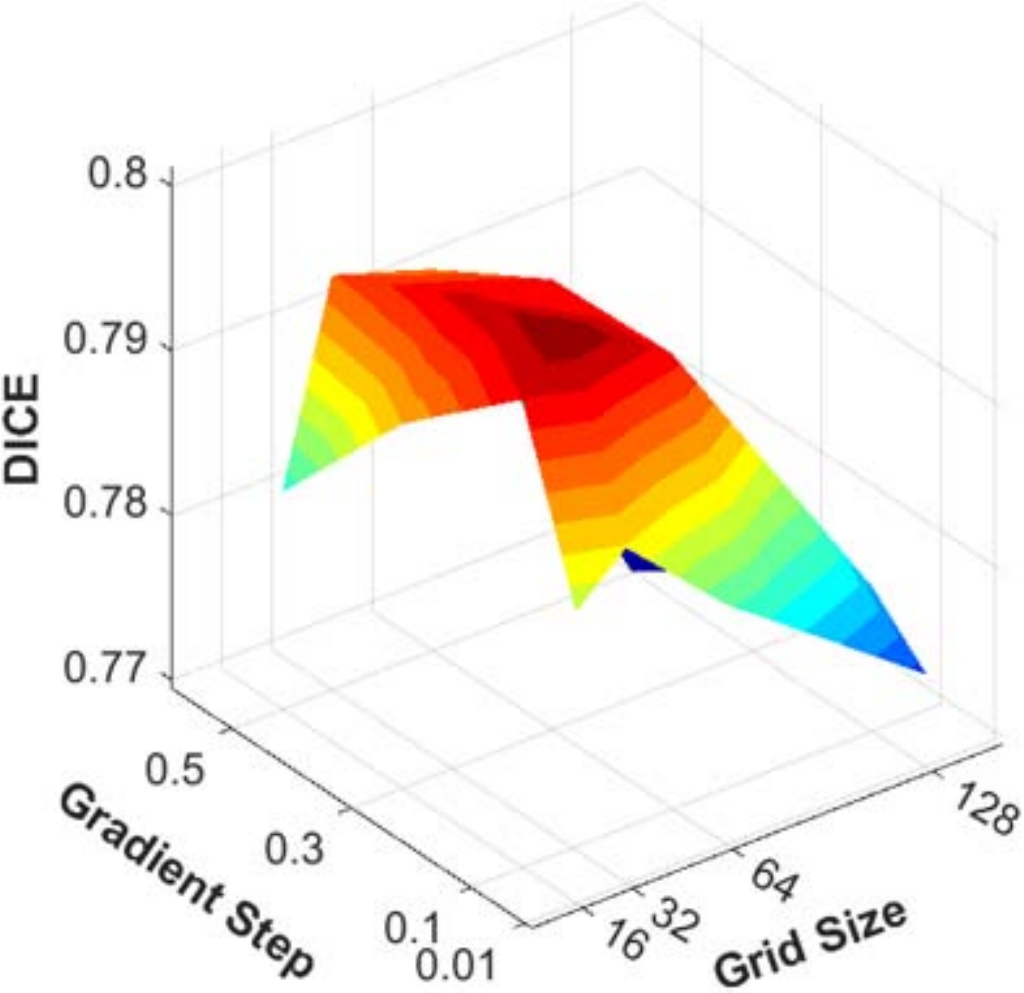}\\
(a) 5 HN Patients & (b) 10 HN Patients & (c) 5 Lung Patients & (d) 30 Lung Patients\\
\end{tabular}
\caption{Optimal DIR hyper-parameter selection via hyper-parameter optimization. The optimal hyper-parameter values for input grid size and gradient step were 32 and 0.3 for HN dataset in (a) \& (b) and were 32 and 0.1 for lung dataset in (c) \& (d).}
\label{fig:bayesian}
\end{center}
\end{figure}

This resulting optimal combination of DIR hyper-parameters was used to create the training DVF. Five randomly selected patients from the training sets for each dataset were used in hyper-parameter optimization. We show that increasing the subset of patients for hyper-parameter optimization did not change the optimal DIR hyper-parameters combination in both HN and lung as shown in Figure \ref{fig:bayesian}.

{After deciding the optimal hyper-parameters for the DIR, we chose two additional hyper-parameter combinations near the optimal one to augment DVF dataset for training.}

\subsection{The Seq2Seq deep learning model}
LSTM networks are recurrent network with memory cell units capable of learning long-range dependency in a temporal sequence. Let $T$ be the number of timepoints of interest and $(X_1,X_2,...,X_T)$ be a sequence of input to the recurrent network, where $X_t \in \mathbb{R}^{M\times{N}\times{3}}$ is a vector field (or an image) for a timepoint $t$, and $M\times{N}$ is the dimension of the vector field in the spatial domain. An LSTM unit contains a memory cell with cell status $c_t$, and the input gate $i_t$, a forget gate $f_t$, an output gate $o_t$ and an output state $H_t$. Compared to conventional LSTM, ConvLSTM is capable of modeling 2D spatio-temporal data sequences by replacing LSTM’s multiplication with spatial local convolution. 
The operation of a ConvLSTM unit is defined by the following equations: 
\cite{kim2017deeprain,mukherjee2019predicting,zhang2019spatio}:

The first step in ConvLSTM is to decide what information to be retained from the cell state by examining the hidden state from the previous timepoint and the input of the current timepoint through the forget gate, $f_t$.
$$f_t=\sigma(W_{f}*[H_{t-1}, X_{t}]+b_{f})$$
And then, the input gate, $i_t$ decides which new features will be stored and a new candidate value, $\tilde{C}_t$, is formed.  
$$i_t=\sigma(W_{i}*[H_{t-1}, X_{t}]+b_{i})$$
$$\tilde{C}_{t}=tanh(W_{c}*[H_{t-1}, X_{t}]+b_{c})$$
Next the new cell state is updated with the new candidate value, $\tilde{C}_t$ and the previous cell state $C_{t-1}$, regulated by the forget and input gates.
$$C_{t} = f_{t} \odot C_{t-1}+i_{t}\odot \tilde{C}_t$$
Finally, the ConvLSTM decides what features are going to pass through the output gate, $o_t$.
$$o_{t} = \sigma(W_{o}*[H_{t-1}, X_{t}]+b_{o})$$
$$H_{t} = o_{t} \odot tanh(C_{t})$$

where $\sigma$ is the sigmoid activation function, $*$ is the convolution operator, and $\odot$ is the Hadamard product. The input $X_{t}$, cell status $C_{t}$, hidden states $H_{t}$, forget gate $f_{t}$, input gate $i_{t}$, input-modulation gate $\tilde{C}_t$, and output gate $o_{t}$ are all 3D tensor with the dimensions of $[\textrm{Width}\times \textrm{Height} \times \textrm{Features}]$. {We used  each slice's 3-axis DVF $\in \mathbb{R}^{M\times{N}\times{3}}$ for inferences to predict corresponded slice's future 3-axis DVF. Through this model, we can predict each slice's 3- axis deformation, and stacked each slice's deformation to generate future 3D volume 3-axis deformation.}

\begin{figure*}[h!]
    \center
    \includegraphics[width=\textwidth]{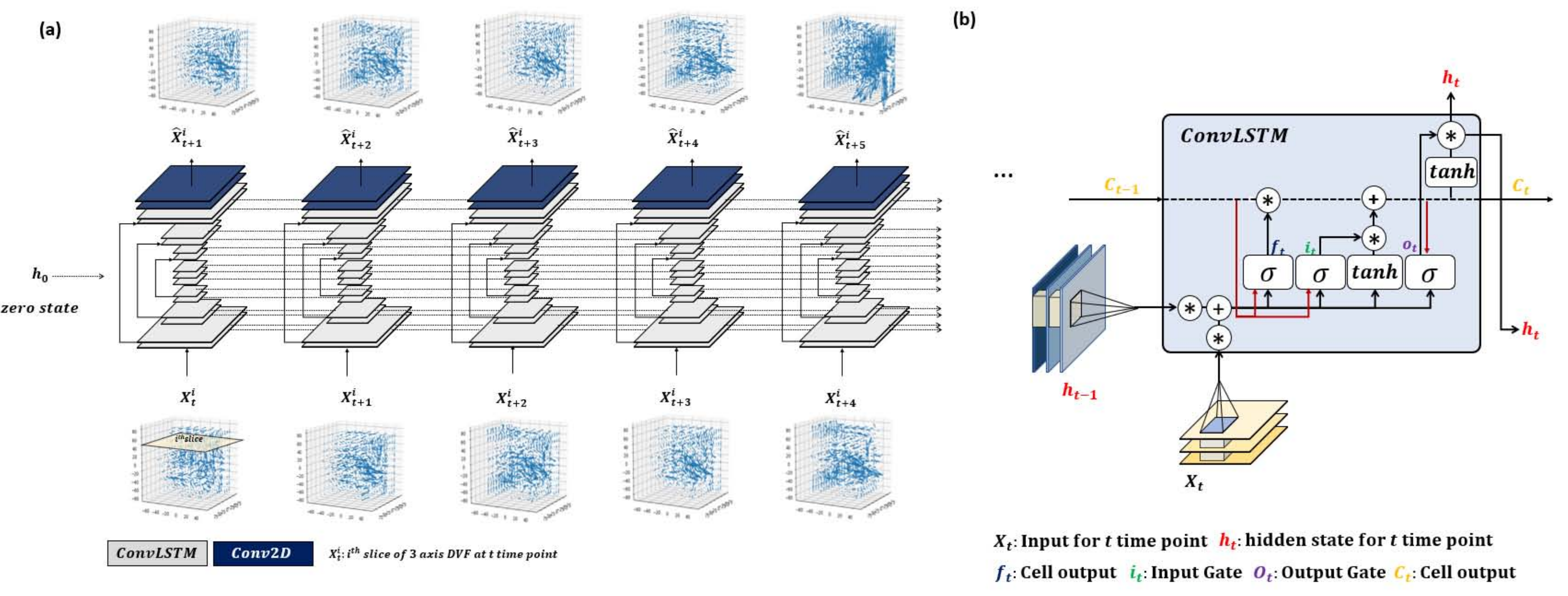}
    \caption{(a) Pipeline of proposed ConvLSTM model, (b) Schematic images for multi resolution ConvLSTM block. Max pooling and upsampling layers are used for multi-resolution longitudinal features analysis.}
    \label{fig:pipeline}
\end{figure*}

Figure \ref{fig:pipeline}(a) is a schematic image of the training process by our deep learning architecture, (b) is ConvLSTM block. 

We built our recurrent network from a recurrent block containing an encoder-decoder sub-network similar to U-Net.  Unlike \cite{zhang2019spatio} in which ConvLSTM is only used in the bottleneck layer, we replaced each convolution layer in the encoder and decoder with ConvLSTM to learn the encoding and decoding dependency from previous timepoints at different resolutions. The encoder consists of three ConvLSTM and two max-pooling layers; the decoder includes three ConvLSTM layers, two upsampling, and two convolution layers. We maintained the long skip connections crossing the encoder and the decoder at the same level of resolution such that fine-grained details can be recovered in the prediction. At the end of the decoder layers, there are two convolution layers to generate the final prediction of future sequential volumetric DVFs.

The recurrent block predicts the DVF of the next timepoint, $\hat{X}_{t+1}$. We defined the network training loss as follows:

$$ Loss = \frac{1}{T-k}\sum^{T-1}_{t=k}{LogCosh(X_{t+1},\hat{X}_{t+1})} $$ and
$$ LogCosh(x, y) = \frac{1}{M\times{}N}\sum^{M\times{N}}_{i=1}{log(cosh(|x_{i}-y_{i}|))}$$

Log-Cosh loss works like L2 for small differences and as L1 for large differences. This means that the loss will not be so strongly affected by the occasional highly incorrect predictions \cite{jadon2020survey}. 

During the training, we used all available timepoints. At inference, the input could be any sequence of previous $K$ timpoints, $1,...,K$, $K<T$, and the network will predict the future sequence of timepoints $K+1,...,T$.
The proposed network has the following advantages: (1) it is possible to predict the future anatomical shape changes at a relatively earlier timepoint during the RT so that, if necessary, the treatment plan could be adapted to the predicted anatomical change in the early phase of treatment, and (2) it can be applied to different input/output sequences of timepoints to update the predictions on a rolling basis as more longitude images are acquired.

To compare with image-based prediction\cite{zhang2019spatio}, in addition to DVF, we trained the proposed network with sequences of $X_t \in \mathbb{R}^{M\times{N}\times{2}}$, where the 2 channels are the image and its anatomical segmentation. We used Adaptive moment estimation (Adam) for optimizer in training both DVF based and image-based networks and learning rate $\alpha$, $\beta 1$, $\beta 2$, and $\epsilon$ in Adam were 0,001, 0.9, 0.999, and 0.1, respectively. 

We used the DVF representation for predicting future anatomical deformations due to its inherent nature in defining relationships with respect to the reference image (planning/pre-treatment images). Representing this relationship is also a clinical requirement since all contouring and dose planning is done on the planning/pre-treatment images and the later timepoints are normally just used for patient setup. We evaluated our model on both image- and DVF-based predictions and showed superior performance with DVF representation. 
{The workstation that we used had Intel® Xeon® Silver 4110 CPU @ 2.10 GHz, 96 GB RAM, and NVIDIA 2080Ti graphic card.}

\subsection{Evaluation}
\subsubsection{Jacobian evaluation}

The purpose of the Jacobian evaluation is to compare the deformation between the DIR and the predicted DVF. For the Jacobian calculation, a 3×3 Jacobian matrix $j$ and its determinant $J$ were calculated from the DVF $\vec{u}$ at every voxel:

$$j(u) = \begin{bmatrix}
\frac{\partial u_x}{\partial x} & \frac{\partial u_x}{\partial y} & \frac{\partial u_x}{\partial z}  \\ \frac{\partial u_y}{\partial x} & \frac{\partial u_y}{\partial y} & \frac{\partial u_y}{\partial z} \\ \frac{\partial u_z}{\partial x } & \frac{\partial u_z}{\partial y} & \frac{\partial u_z}{\partial z}
\end{bmatrix}$$

$$ J(u) = Det(j(u))$$

$j$ is the first derivative of the DVF and is calculated at every voxel to produce a map of $J$ \cite{riyahi2018quantifying,chung2001unified,fuentes2015morphometry}.

\begin{figure}[h!]
\begin{center}
\footnotesize
\setlength{\tabcolsep}{3pt}
\includegraphics[width=0.46\textwidth]{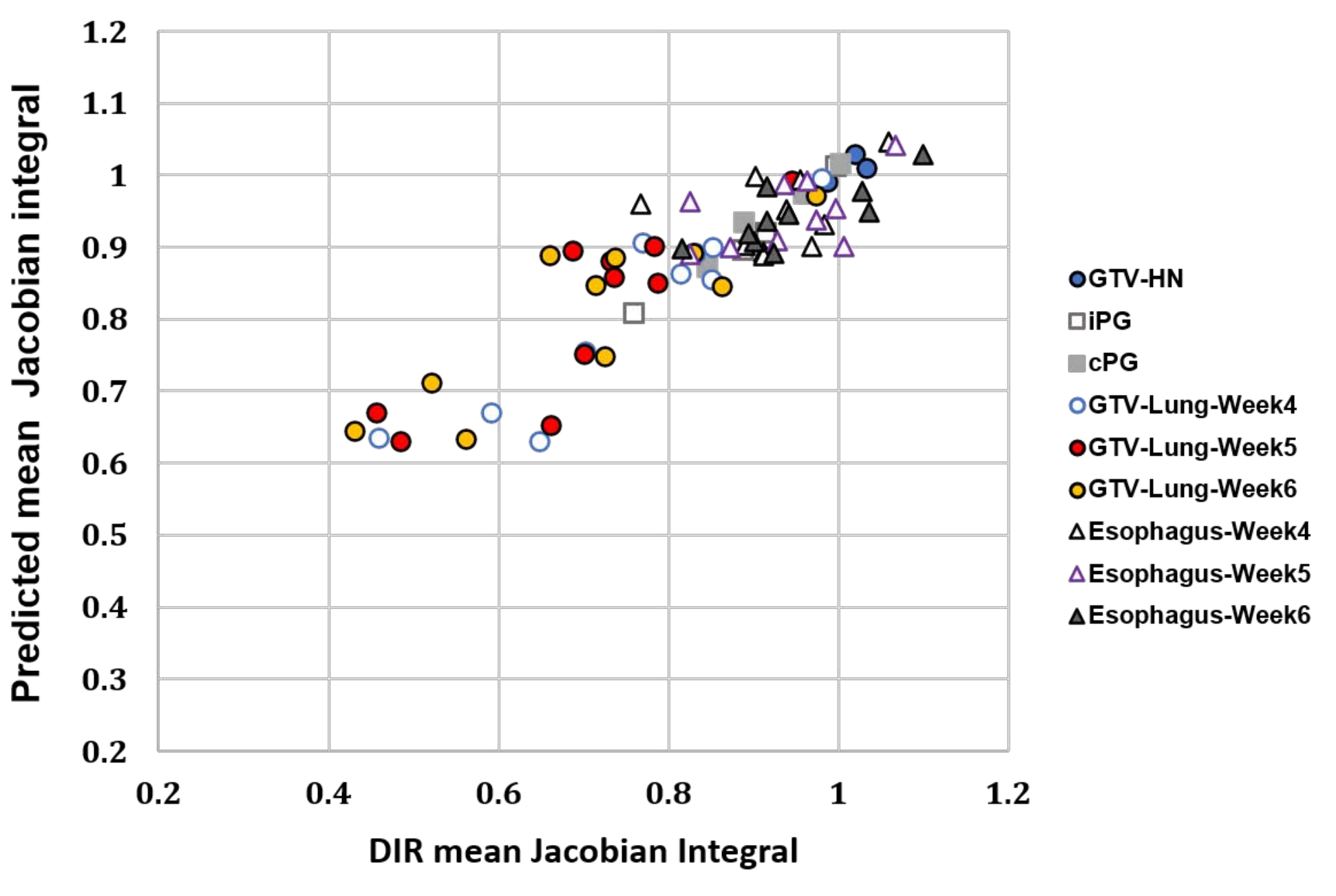}
\caption{Jacobian evaluation on HN and lung datasets with prediction and DIR-derived DVFs. The Jacobian integral for predicted and DIR DVFs showed high correlation, $R^2=0.85$.}
\label{fig:jacobian}
\end{center}
\end{figure}

\begin{figure}[ht!]
\begin{center}
\footnotesize
\setlength{\tabcolsep}{1pt}
\includegraphics[width=1\textwidth]{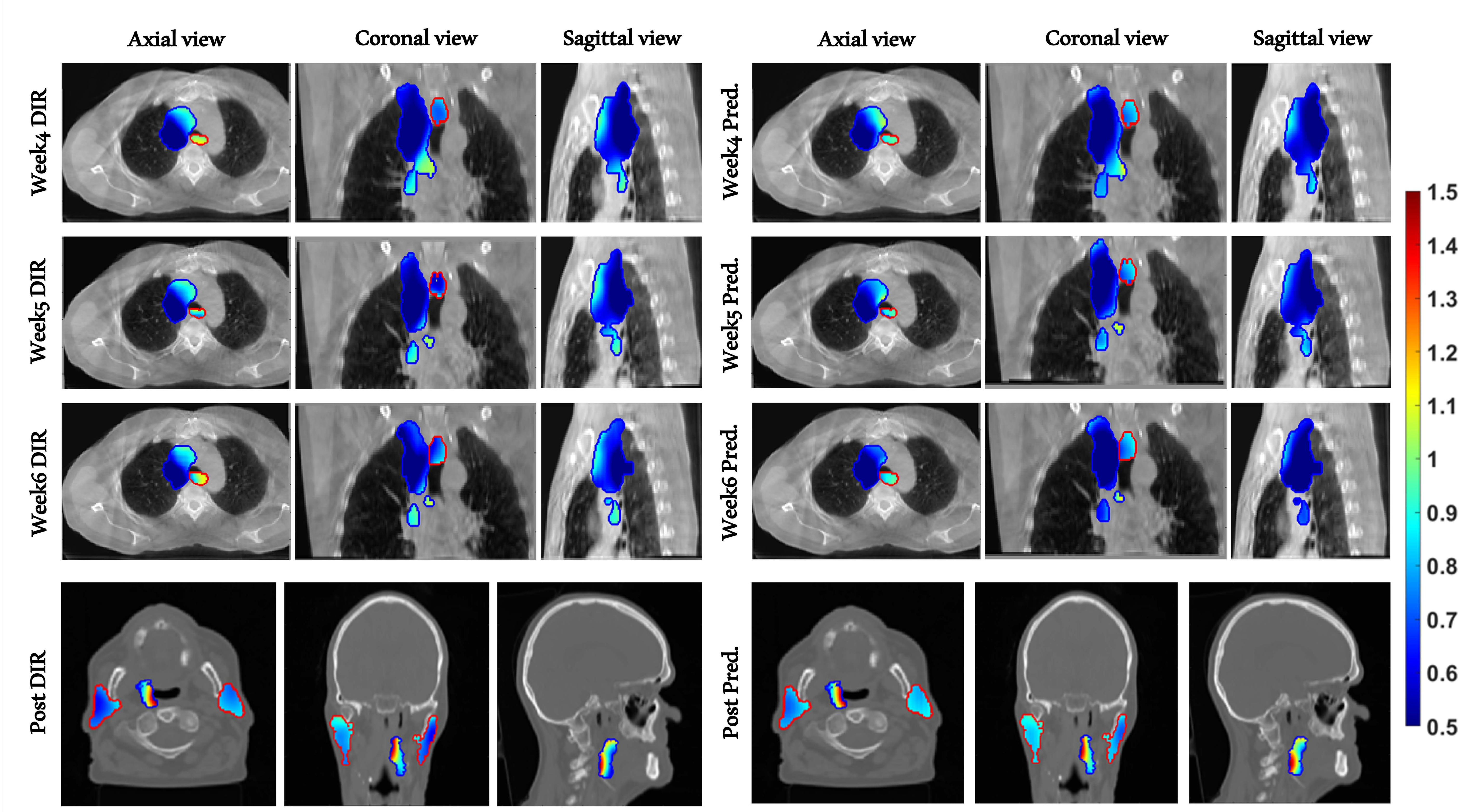}
\caption{Comparison of DIR and predicted Jacobian maps for the lung example in Figure 1 and HN example in Figure 2. $J<1$ indicates shrinkage, $J>1$ expansion, and $J=1$ is no change.}
\label{fig:jacobian_map}
\end{center}
\end{figure}

The Jacobian map indicates the volumetric ratio of an object before and after transformation. $J>1$ means volume expansion and $J<1$ corresponds to volume shrinkage. As shown in Figures \ref{fig:jacobian} and \ref{fig:jacobian_map}, we also quantitatively compared $J$ from DIR-derived DVF and predicted DVF for GTV and parotid glands in HN dataset and for GTV and esophagus in the lung dataset.

\subsubsection{Geometric matching}
The purpose of geometric matching evaluation is to assess the ability of the proposed approach to predict anatomical deformation between initial images and future images. In the case of image-based prediction, we directly measured the DICE coefficient, and Average Hausdorff distance $(d_{AVD})$ between manual contours and predicted contours at a later timepoint. For geometric matching evaluation of DVF-based prediction, we measured the dice coefficient and $d_{AVD}$ of contours warped by predicted DVFs and manual contours. The manual contours were delineated by experienced radiation oncologists. Moreover, we compared the volume of manual and predicted contours for weekly CBCT and for mid-/post-treatment CTs by relative volume difference (RVD) \cite{zhang2019spatio,wong2016pancreatic,roque2017dce,zhang2017personalized,zhang2017convolutional}.
 
$$ DICE  = {2\vert X\cap Y \vert \over \vert X \vert + \vert Y \vert} $$

$$ d_{AVD}(X,Y) = \frac{1}{|X|}\sum_{x\in X} \min_{y\in Y} d(x,y) + \frac{1}{|Y|}\sum_{y\in Y} \min_{x\in X} d(x, y) $$

$$ RVD = {\vert V_x - V_y \vert \over \vert V_x \vert} $$

where $X$ is ground truth and $Y$ is the predicted contours, and $V_x$, $V_y$ is the volume of ground truth and prediction, respectively. The $|X|$ and $|Y|$ are the number of points in $X$ and $Y$, respectively. $d(x,y)$ is Euclidean distance metric between two points $x\in X$ and $y \in Y$
 
\section{Results}
\begin{figure}[ht!]
\begin{center}
\footnotesize
\setlength{\tabcolsep}{1pt}
\includegraphics[width=1\textwidth]{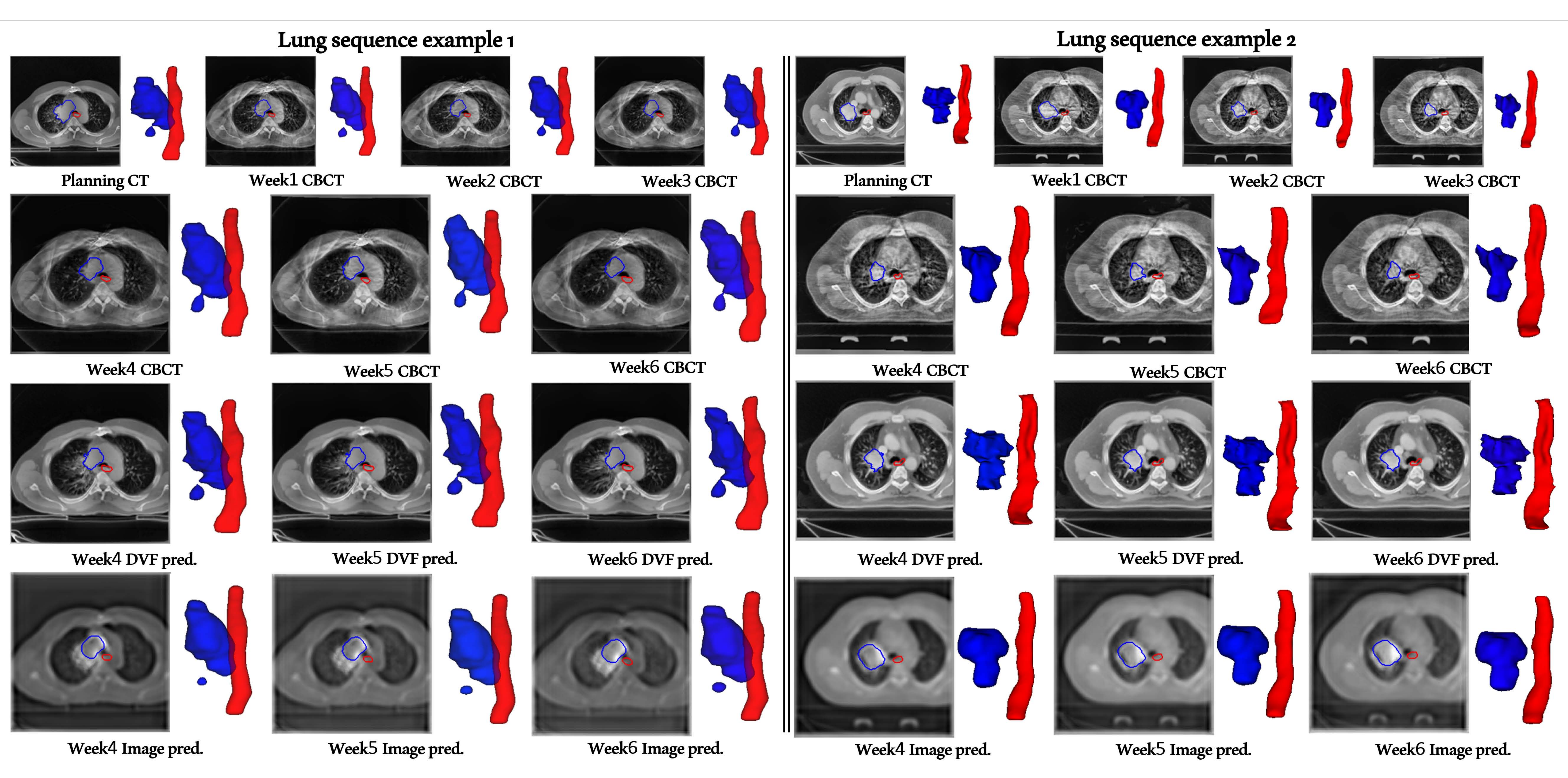}
\caption{DVF- versus image-based sequence prediction of weeks 4, 5 and 6 given the weeks 1, 2 and 3 sequence as input. Note that our DVF representation (unlike image representation) does not contain any high-frequency components and hence, is not affected by the ConvLSTM module, resulting in sharper planning CT deformed week 4, 5 and 6 sequence prediction. Blue contours denote gross tumor volume (GTV) and red denote esophagus (ESO) contours. Deformed planning CT weekly predictions result in artifact/noise-free images as opposed to the CBCT artifact-ridden and blurry predictions with the image representation. \underline{Lung sequence example 1 (Left)}, predicted \textbf{week4 Dice:} DVF-GTV 0.872, Image-GTV 0.89, DVF-ESO 0.73, Image-ESO 0.73; \textbf{week5 Dice:} DVF-GTV 0.83, Image-GTV 0.82, DVF-ESO 0.82, Image-ESO 0.81; \textbf{week6 Dice}: DVF-GTV 0.83, Image-GTV 0.79, DVF-ESO 0.76, Image-ESO 0.76. 
\underline{Lung sequence example 2 (Right)}, predicted \textbf{week4 Dice:} DVF-GTV 0.77, Image-GTV 0.70, DVF-ESO 0.77, Image-ESO 0.80; \textbf{week5 Dice:} DVF-GTV 0.73, Image-GTV 0.65, DVF-ESO 0.80, Image-ESO 0.81; \textbf{week6 Dice}: DVF-GTV 0.64, Image-GTV 0.59, DVF-ESO 0.82, Image-ESO 0.80.}
\label{fig:lung_examples}
\end{center}
\end{figure}

Figure \ref{fig:lung_examples} shows two examples of Seq2Seq prediction results for lung GTV (blue) and esophagus (red). Both lung sequence examples demonstrate weeks 4--6 sequence prediction results using DVF and image representations in our Seq2Seq model. The lung sequence example 1 is from GTV undergoing inflammation (increase in volume) from weeks 1--3 but at the end exhibits overall 20\% shrinkage from the start of the treatment. 

\begin{figure}[ht!]
\begin{center}
\footnotesize
\setlength{\tabcolsep}{1pt}
\includegraphics[width=1\textwidth]{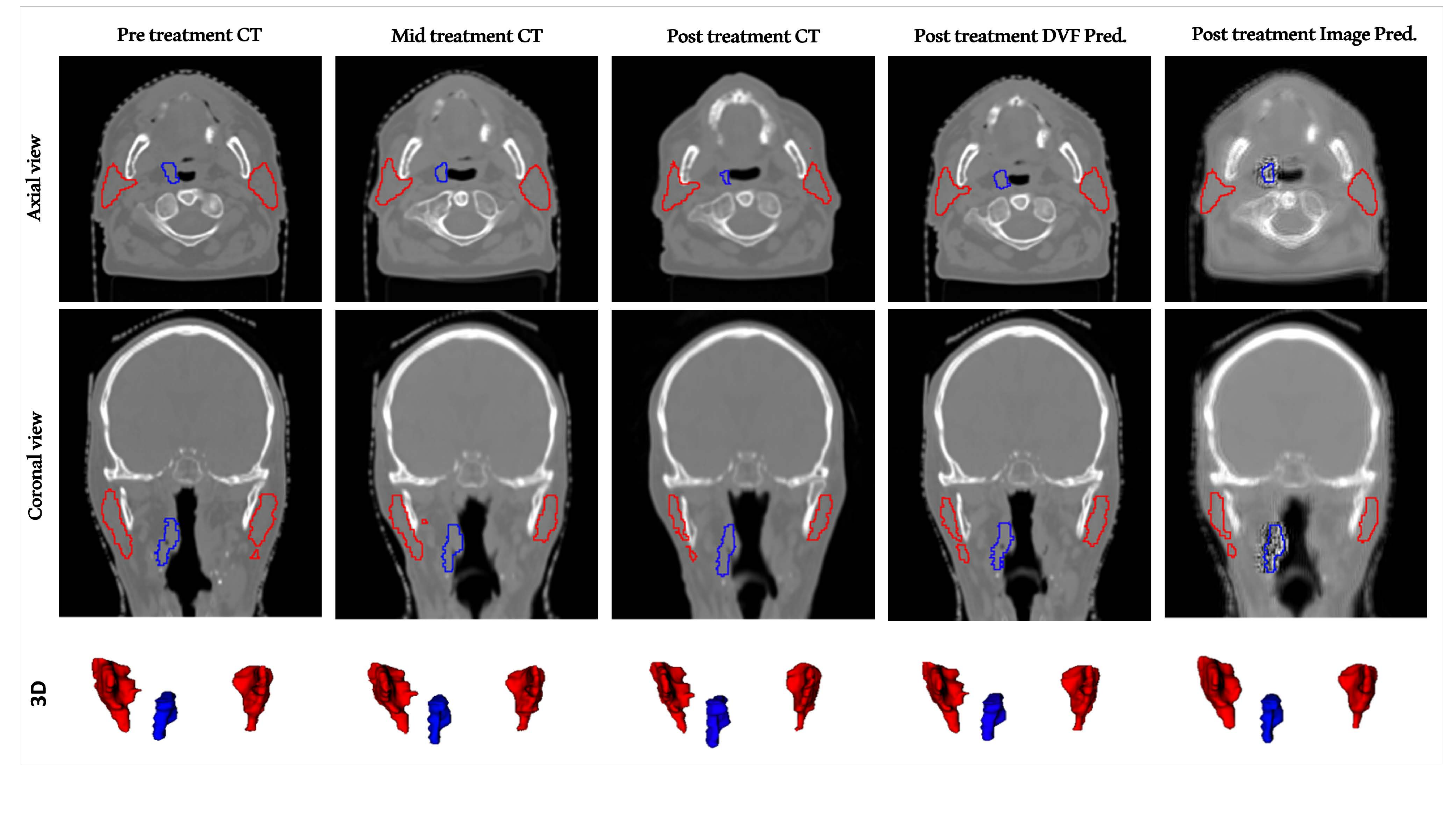}
\caption{HN radiotherapy dataset example illustrating DVF- versus image-based \cite{zhang2019spatio} prediction results for post-treatment CT given pre- and mid-treatment CTs as input sequence. Blue contours represents GTV and red denote parotid gland, ipsilateral (iPG) and contralateral (cPG) sides. Dice, average Hausdorff (AvHD) and relative volume difference (RVD) for \textbf{image prediction} are as follows, GTV-DICE: 0.67, GTV-AvHD: 0.62, GTV-RVD: 26.8\%, iPG-DICE: 0.59, iPG-AvHD: 1.12, iPG-RVD: 25.8\%, cPG-DICE: 0.80, cPG-AvHD: 0.33, cPG-RVD: 17.9\%. For \textbf{DVF prediction}, GTV-DICE: 0.75, GTV-AvHD: 0.39, GTV-RVD: 7.9\%, iPG-DICE: 0.72, iPG-AvHD: 0.48, iPG-RVD: 24.1\%, cPG-DICE: 0.83, cPG-AvHD: 0.24, cPG-RVD: 16.2\%.}
\label{fig:hn_examples}
\end{center}
\end{figure}

\begin{table*}[h]
\small
\caption{Metrics (Dice, Average Hausdorff, Relative Volume Difference) evaluation for HN and lung datasets.}
\label{table:Metrics}
\setlength\tabcolsep{4pt}
\centering
\begin{tabular}{|l|c|c|c|c|c|c|}
\hline
& Image Pred. &  DVF Pred. & Image Pred. & DVF Pred. & Image Pred. &  DVF Pred.\\
\hline
\textbf{Structures}  &   \multicolumn{2}{c|}{\textbf{DICE $\uparrow$}} &  \multicolumn{2}{c|}{\textbf{$d_{AVD}(mm) \downarrow$}} &   \multicolumn{2}{c|}{\textbf{RVD (\%)$\downarrow$}}\\
\hline
\multicolumn{7}{|c|}{\textbf{HN post-treatment image predictions without skip connections}}\\
\hline
GTV-HN & $0.82\pm0.11$ & $0.84\pm0.11$ & $0.33\pm0.24$ & $0.22\pm0.25$ & $12.6\pm10.6$ & $4.6\pm2.6$\\
iPG-HN & $0.76\pm0.08$ & $0.78\pm0.08$ & $0.52\pm0.25$ & $0.33\pm0.27$ & $13.7\pm14.1$ & $10.9\pm9.2$\\
cPG-HN & $0.81\pm0.05$ & $0.84\pm0.05$ & $0.31\pm0.13$ & $0.22\pm0.18$ & $6.4\pm4.6$ & $9.1\pm9.5$\\
\hline
\multicolumn{7}{|c|}{\textbf{HN post-treatment image predictions with skip connections}}\\
\hline
GTV-HN & $0.84\pm0.10$ & $0.87\pm0.07$ & $0.28\pm0.21$ & $0.21\pm0.13$ & $11.9\pm14.1$ & $4.5\pm2.5$\\
iPG-HN & $0.76\pm0.11$ & $0.81\pm0.06$ & $0.52\pm0.34$ & $0.32\pm0.13$ & $13.3\pm14.5$ & $8.1\pm8.2$\\
cPG-HN & $0.84\pm0.04$ & $0.86\pm0.02$ & $0.28\pm0.21$ & $0.21\pm0.13$ & $5.7\pm6.0$ & $4.5\pm6.5$\\
\hline
\multicolumn{7}{|c|}{\textbf{Lung Weeks 4--6 prediction (without skip connections)}}\\
\hline
GTV-Lung-Week4 & $0.81\pm0.09$ & $0.82\pm0.10$ & $0.31\pm0.19$ & $0.28\pm0.20$ & $24.2\pm31.5$ & $19.1\pm25.5$\\
GTV-Lung-Week5 & $0.79\pm0.07$ & $0.81\pm0.08$ & $0.36\pm0.15$ & $0.29\pm0.17$ & $34.7\pm27.6$ & $22.3\pm20.2$\\
GTV-Lung-Week6 & $0.77\pm0.10$ & $0.79\pm0.11$ & $0.42\pm0.22$ & $0.33\pm0.22$ & $48.0\pm45.8$ & $26.7\pm34.7$\\
Eso-Lung-Week4 & $0.73\pm0.06$ & $0.75\pm0.03$ & $0.36\pm0.11$ & $0.29\pm0.07$ & $27.9\pm23.6$ & $27.9\pm12.4$\\
Eso-Lung-Week5 & $0.73\pm0.06$ & $0.76\pm0.05$ & $0.42\pm0.17$ & $0.36\pm0.16$ & $26.7\pm19.8$ & $27.4\pm14.6$\\
Eso-Lung-Week6 & $0.73\pm0.05$ & $0.77\pm0.04$ & $0.41\pm0.17$ & $0.30\pm0.08$ & $28.9\pm28.4$ & $19.1\pm17.1$\\
\hline
\multicolumn{7}{|c|}{\textbf{Lung Weeks 4--6 prediction (with skip connections)}}\\
\hline
GTV-Lung-Week4 & $0.81\pm0.09$ & $0.83\pm0.09$ & $0.31\pm0.24$ & $0.26\pm0.19$ & $21.9\pm28.7$ & $15.5\pm22.5$\\
GTV-Lung-Week5 & $0.80\pm0.07$ & $0.82\pm0.08$ & $0.36\pm0.23$ & $0.29\pm0.18$ & $21.8\pm23.7$ & $17.8\pm16.2$\\
GTV-Lung-Week6 & $0.79\pm0.09$ & $0.81\pm0.10$ & $0.40\pm0.20$ & $0.31\pm0.21$ & $27.8\pm31.4$ & $22.2\pm27.1$\\
Eso-Lung-Week4 & $0.76\pm0.04$ & $0.77\pm0.03$ & $0.29\pm0.18$ & $0.29\pm0.08$ & $24.1\pm6.8$ & $23.6\pm11.6$\\
Eso-Lung-Week5 & $0.76\pm0.07$ & $0.77\pm0.05$ & $0.35\pm0.15$ & $0.35\pm0.16$ & $24.0\pm14.8$ & $22.7\pm13.8$\\
Eso-Lung-Week6 & $0.77\pm0.05$ & $0.77\pm0.03$ & $0.29\pm0.08$ & $0.29\pm0.08$ & $23.5\pm14.7$ & $15.0\pm15.4$\\
\hline
\multicolumn{7}{|c|}{\textbf{Lung 3--6 prediction (with skip connections)}}\\
\hline
GTV-Lung-Week3 & $0.83\pm0.10$ & $0.84\pm0.13$ & $0.26\pm0.24$ & $0.24\pm0.17$ & $20.4\pm29.4$ & $19.1\pm12.0$\\
GTV-Lung-Week4 & $0.80\pm0.12$ & $0.81\pm0.11$ & $0.34\pm0.27$ & $0.31\pm0.23$ & $27.1\pm37.5$ & $26.7\pm42.5$\\
GTV-Lung-Week5 & $0.76\pm0.11$ & $0.80\pm0.09$ & $0.37\pm0.24$ & $0.31\pm0.21$ & $33.3\pm42.6$ & $31.1\pm31.3$\\
GTV-Lung-Week6 & $0.76\pm0.14$ & $0.78\pm0.12$ & $0.45\pm0.34$ & $0.36\pm0.27$ & $44.2\pm58.6$ & $36.1\pm48.7$\\
Eso-Lung-Week3 & $0.78\pm0.03$ & $0.78\pm0.04$ & $0.30\pm0.06$ & $0.30\pm0.09$ & $27.8\pm20.3$ & $26.8\pm18.3$\\
Eso-Lung-Week4 & $0.77\pm0.04$ & $0.77\pm0.03$ & $0.29\pm0.06$ & $0.29\pm0.06$ & $27.9\pm10.4$ & $26.4\pm10.1$\\
Eso-Lung-Week5 & $0.76\pm0.05$ & $0.76\pm0.06$ & $0.36\pm0.13$ & $0.36\pm0.14$ & $27.4\pm14.6$ & $25.9\pm14.4$\\
Eso-Lung-Week6 & $0.77\pm0.04$ & $0.77\pm0.03$ & $0.30\pm0.07$ & $0.30\pm0.05$ & $24.3\pm14.7$ & $17.1\pm16.0$\\
\hline
\end{tabular}
\end{table*}

Since our training data contained 18 cases (Figure \ref{fig:volume_parallel}) from the inflammation category, even though the lung example 1 input weeks 1--3 sequence volume is monotonically increasing, our DVF model is correctly able to predict the eventual shrinkage with sharper boundaries whereas the image-based prediction gets thrown off in later weeks (and produces significantly blurry results). Lung sequence example 2 belongs to the typical monotonically decreasing volume category and exhibits overall 50\% shrinkage. The training/validation data contained 35 cases from this category (Figure \ref{fig:volume_parallel}) and the Seq2Seq model shows good GTV and esophagus prediction performance on this category (Figure \ref{fig:lung_examples} and Table \ref{table:Category_results}). 

\begin{table*}[ht]
\caption{Lung GTV prediction metrics evaluation for inflammation (1) and monotonically decreasing (2) volume categories.}
\label{table:Category_results}
\setlength\tabcolsep{4pt}

\centering
\begin{tabular}{|l|c|c|c|c|c|c|}
\hline
& Image Pred. &  DVF Pred. & Image Pred. & DVF Pred. & Image Pred. &  DVF Pred.\\
\hline
Structures  &   \multicolumn{2}{c|}{\textbf{DICE $\uparrow$}} &  \multicolumn{2}{c|}{\textbf{ $d_{AVD}(mm) \downarrow$}} &   \multicolumn{2}{c|}{\textbf{RVD (\%) $\downarrow$}}\\
\hline
Category1-Week4 & $0.85\pm0.11$ & $0.85\pm0.12$ & $0.26\pm0.28$ & $0.21\pm0.20$ & $4.8\pm1.5$ & $4.7\pm3.4$\\
Category1-Week5 & $0.85\pm0.10$ & $0.86\pm0.10$ & $0.28\pm0.33$ & $0.20\pm0.15$ & $8.6\pm5.6$ & $8.1\pm8.1$\\
Category1-Week6 & $0.86\pm0.06$ & $0.87\pm0.08$ & $0.21\pm0.20$ & $0.19\pm0.12$ & $8.8\pm10.6$ & $8.6\pm1.7$\\
Category2-Week4 & $0.79\pm0.11$ & $0.82\pm0.09$ & $0.35\pm0.23$ & $0.27\pm0.19$ & $28.1\pm32.4$ & $20.1\pm25.9$\\
Category2-Week5 & $0.76\pm0.10$ & $0.80\pm0.07$ & $0.39\pm0.20$ & $0.29\pm0.19$ & $27.6\pm26.5$ & $21.9\pm17.5$\\
Category2-Week6 & $0.73\pm0.13$ & $0.81\pm0.10$ & $0.48\pm0.30$ & $0.35\pm0.21$ & $35.9\pm34.5$ & $28.0\pm31.2$\\
\hline
\end{tabular}
\end{table*}

Note that the DVF-based prediction deforms planning CT images and hence produces sharper noise-/artifact-free predictions corresponding to the weekly CBCTs, showing another advantage of using DVF rather than image representation (where artifacts/noise in CBCT images can potentially throw off predictions). 

\begin{figure}[ht!]
\begin{center}
\footnotesize
\setlength{\tabcolsep}{1pt}
\includegraphics[width=0.8\textwidth]{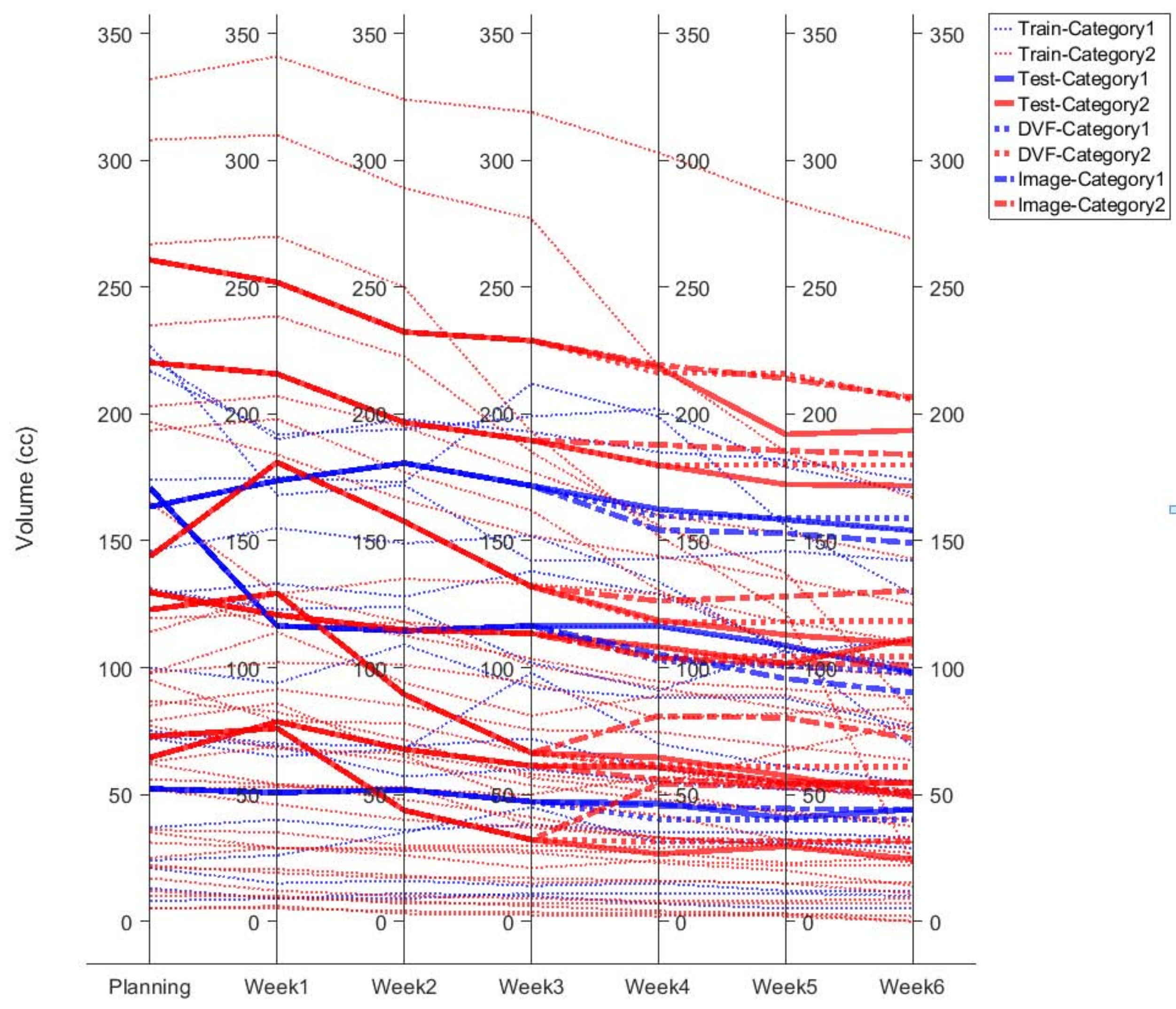}
\caption{The absolute volume of two different lung GTV categories: Category 1 (blue) with inflammation leading to non-monotonic volume changes through the course of treatment and Category 2 (red) with monotonically decreasing volume. There were 18 category-1 and 35 category-2 cases in training/validation and 3 category-1 and 7 category-2 cases in testing.}
\label{fig:volume_parallel}
\end{center}
\end{figure}

Detailed quantitative evaluations on inflammation and monotonically decreasing volume categories for lung dataset is given in Figure \ref{fig:volume_parallel} and Table \ref{table:Category_results}. There were 18 category 1 (inflammation) and 35 category 2 (monotonic volume decrease) cases in training/validation sets, and 3 category 1 and 7 category 2 cases in the test set. The presented DVF Seq2Seq model was shown to accurately predict future GTV volumes even if there were volume fluctuations in the input sequence. Image-based prediction overestimated GTV volumes in the inflammation category cases by over 30\% $RVD$. As compared to modest overall GTV shrinkage in category 1, the monotonic decreasing category exhibited larger shrinkage ranges, some as high as 50\%. Understandably, the prediction accuracy was relatively lower in category 2 than category 1 but the overall prediction results for both categories with DVF representation are clinically reasonable (over 0.8 $DICE$).

Figure \ref{fig:hn_examples} shows the post-treatment prediction results for a HN patient given pre- and mid-treatment sequence input. We investigated GTV (blue) and the PG (red) for HN evaluation. Both GTV and PG shrunk during the RT and both image- and DVF-based predictions performed well though DVF-based prediction achieved 0.75 GTV $DICE$, approximately 13\% better than image-based prediction. The PG normally exhibits the largest volume change among OARs during HN RT and the proposed DVF prediction obtained 0.72 DICE for iPG and 0.83 for cPG, approximately 23\% and 4.5\% better than image-based iPG and cPG predictions, respectively. Not only does DVF prediction achieves better accuracy, but the resulting images are also clear/sharp, overcoming the major bottleneck with the image-based prediction. 

Since ConvLSTM models suffer from mode-averaging and limited long-term dependencies between sequence elements, we overcame the mode-averaging issue via DVF representation and limited long-term dependency issue via skip connections by directly propagating extracted multi-resolution longitudinal features from encoder to decoder ConvLSTM block. The metric evaluation for predictions via image vs DVF representation and with vs without skip connections are reported in Table \ref{table:Metrics} which shows DICE, Average Hausdorff, and Relative volume difference evaluation for GTVs and OARs in both lung and HN datasets. The DVF and skip connection combination outperforms image and with/without skip connection combination. Since our model has a flexible input/output sequence structure, we also report weeks 1--2 and weeks 1--3 sequence input predictions. The prediction results from weeks 1--2 were slightly lower than weeks 1--3, but the performance was still clinically valuable especially since DVF prediction method maintained relatively good results even though the input data for future prediction was insufficient. {Further, if we input sequences up to week 4 or week 5, we were able to obtain week 6 prediction with DICE of $0.83\pm0.08$ and $0.84\pm0.08$, respectively. From these results, we demonstrated the advantage of our proposed seq2seq model in handling various lengths of input sequence on a rolling basis as the timepoints are acquired to improve prediction accuracy.} 

{The mean and standard deviation of GTV volume change from planning CT to week 6 CBCT were 111.2$\pm$81.7, 106.9$\pm$79.9, 101.4$\pm$76.9, 95.6$\pm$74.6, 88.7$\pm$68.7, 81.6$\pm$62.6 and 75.3$\pm$60.1, respectively. The testing data’s GTV volume from week4 to 6 was (110.2$\pm$62.3, 102.9$\pm$57.3, 101.1$\pm$58.2), and the volume of predicted GTV by DVF based method was (120.3$\pm$70.4, 118.8$\pm$71.2, 108.4$\pm$61.5), and image-based method was (131.1$\pm$65, 136.8$\pm$64.9, 134.1$\pm$64.5), respectively. Figure \ref {fig:volume_parallel} is the volume change tendencies of GTV and we can confirm that the decreased tendencies of ground truth GTV and predicted GTV volume, especially predicted by DVF based method, were similar to each other.}

{And we also confirmed that the prediction performance was slightly degraded as predicting timepoint was increased, and this phenomenon was derived from error accumulation of previous predictions}

\section{Discussion}
During the RT, the patient's anatomy significantly changes due to weight loss and radiation-induced biological effects. These changes can increase uncertainty of radiation dose delivery during the RT. In effect, it is highly desirable to predict patient’s radiation-induced anatomical changes in advance. Recently, longitudinal prediction deep learning models have been proposed for medical imaging which outperform more traditional mathematical models \cite{zhang2019spatio}. This attests to the promising potential for time-series medical image prediction for the RT. Previous studies were evaluated on relatively small datasets with only single timepoint prediction and small region-of-interest (ROI) image patches \cite{zhang2019spatio}.

In this paper, we used three timepoint CT 28 patient HN dataset and six timepoint multimodal CT/CBCT 63 patient lung dataset for developing GTV and OAR prediction. To the best of our knowledge, ours is the largest longitudinal clinical imaging dataset used for developing and evaluating longitudinal sequential prediction. Furthermore, our study has the following strengths: (1) we used the entire 3D field-of-view (FOV) of the imaging dataset rather than the cropped ROI image. Zhang et al. \cite{zhang2019spatio} used only cropped ROI images which resulted in better predictions than would be possible with the complete FOV uncropped images. Their approach though understandable given the unnecessary background information that exists in the whole image which can potentially reduce prediction accuracy, still adds to another variable (cropping margins) in evaluation.

In RT, GTV shape is not the only confounding factor and the surrounding anatomical structures are also an important consideration for delivering a high conformal dose to the target while minimizing radiation dose to the normal tissue. Hence, it is necessary to predict whole CT image anatomical changes during the RT. For our images, we only cropped air and table from the original volume data and used 256×256×256 volume size HN dataset and 180×180×128 volume size Lung dataset which included all the anatomical structures originally scanned. 

(2) We proposed DVF instead of imaging representation for driving the prediction. Since the intensity of a pixel or voxel for the deep learning training is one of the factors that have the greatest influence on the extracted features, deformation vector field is used instead of the image HU values. In DVF, the absolute vector intensity is large in the large deformation regions which can naturally help focus on these regions when training with the DVF representation. Moreover, DVF-based training (with respect to the planning CT) for longitudinal sequential anatomical prediction is more effective since there is no image noise or artifacts involved especially for the lung CBCT cases. The low-dose CBCT modality, generally used for checking patient condition and position weekly during the RT, has significant image noise and artifacts. Thus, training with CBCT imaging may suffer from significant bias from these image noise and artifact. {Using the future DVF prediction, the high-quality planning CT images (along with the contour and planned dose information) can be deformed to the later timepoints thus allowing for re-planning given the same Hounsfield unit of the deformed image as planning CT.} Moreover, the DVF representation per-voxel contains 3 axis direction deformation which aids in 3D spatio-temporal prediction. 

Because we used the DVF for training, the accuracy of DVF is critical. Therefore we used LDDMM based DIR, which is one of the state-of-the-art DIR algorithms. Given that we had manual contours on all the timepoints, we were able to use a fraction of the training data in the hyper-parameter optimization framework to find optimal values for the most sensitive LDDMM hyper-parameters. This resulted in a high-quality DVF training data. 

(3) We used a multi-resolution ConvLSTM based model. The previous Seq2Seq model used ConvLSTM or LSTM block at only high-end features whereas our proposed model uses ConvLSTM block in a multi-resolution block thus allowing for deeper investigation of the sequential features. For example, the week 6 prediction of lung GTV with the multi-resolution ConvLSTM model was $0.81\pm0.10$ average DICE, but the prediction result with the model that has ConvLSTM block only in the middle part was $0.79\pm0.09$. 

(4) {Lastly our study comprises of larger and longer real clinical sequence datasets than reported in previous works. We used 6-timepoint sequences from 63 non-small cell lung cancer patients. In comparison, Zhang et al.\cite{zhang2019spatio} only predicted a single timepoint given two previous time points. Furthermore, we studied multiple prediction results from different length input/output sequences. Given the requirement for delivering high conformal dose to the target structure while minimizing OAR dose, our model provides the flexibility to make longitudinal predictions on a rolling basis as more weekly images are acquired.}

\begin{figure}[ht!]
\begin{center}
\footnotesize
\setlength{\tabcolsep}{1pt}
\includegraphics[width=0.8\textwidth]{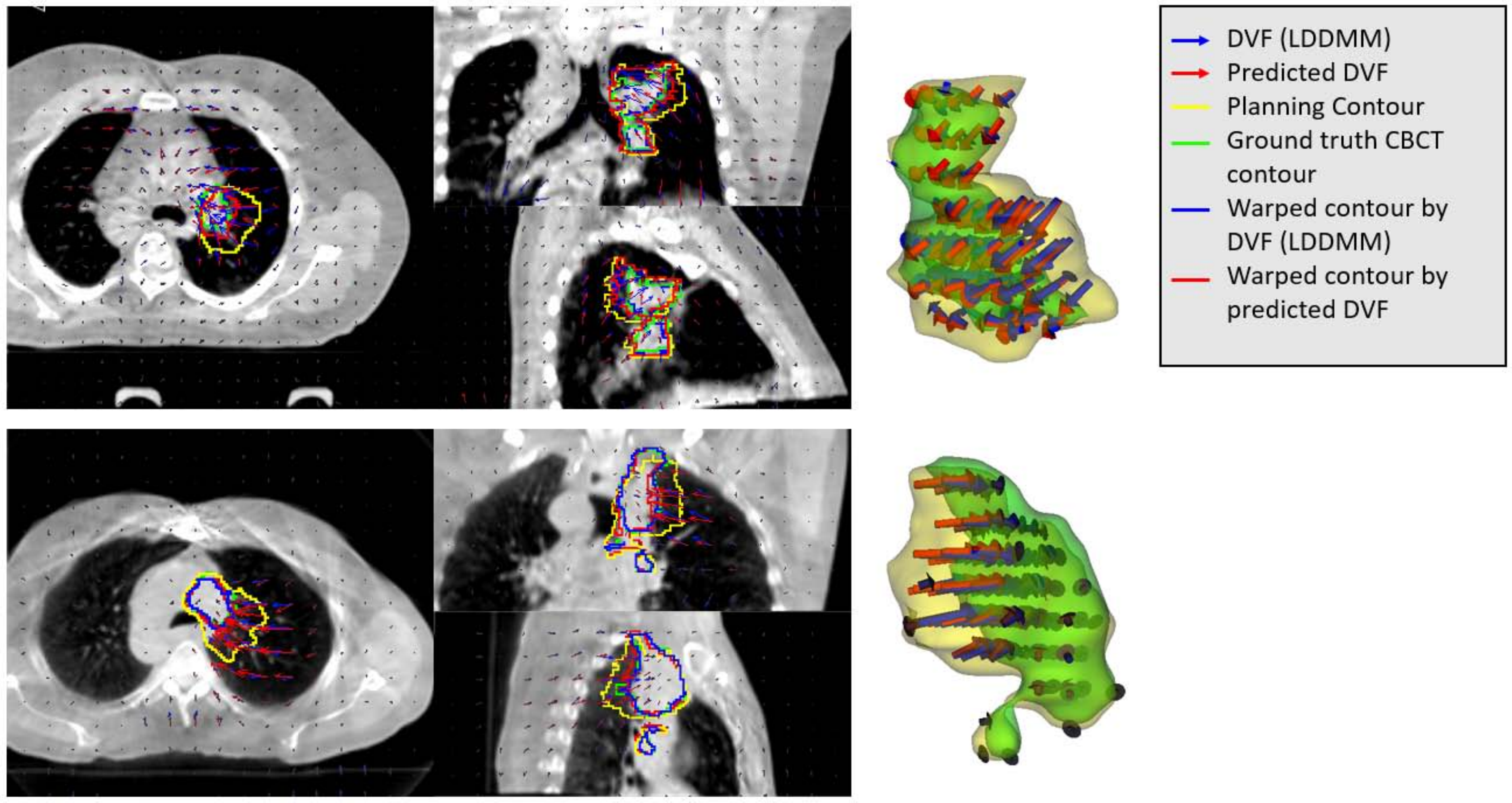}
\caption{{Two examples of DVF, and contour prediction. The background image is week 6 CBCT. The Blue and the red arrow are actual and predicted DVF respectively. Yellow, green, blue, and red contour are planning, Week 6 ground truth, warped GTV by actual DVF and predicted DVF, respectively.}}
\label{fig:figure8}
\end{center}
\end{figure}

\begin{figure}[h]
\begin{center}
\footnotesize
\setlength{\tabcolsep}{1pt}
\includegraphics[width=0.8\textwidth]{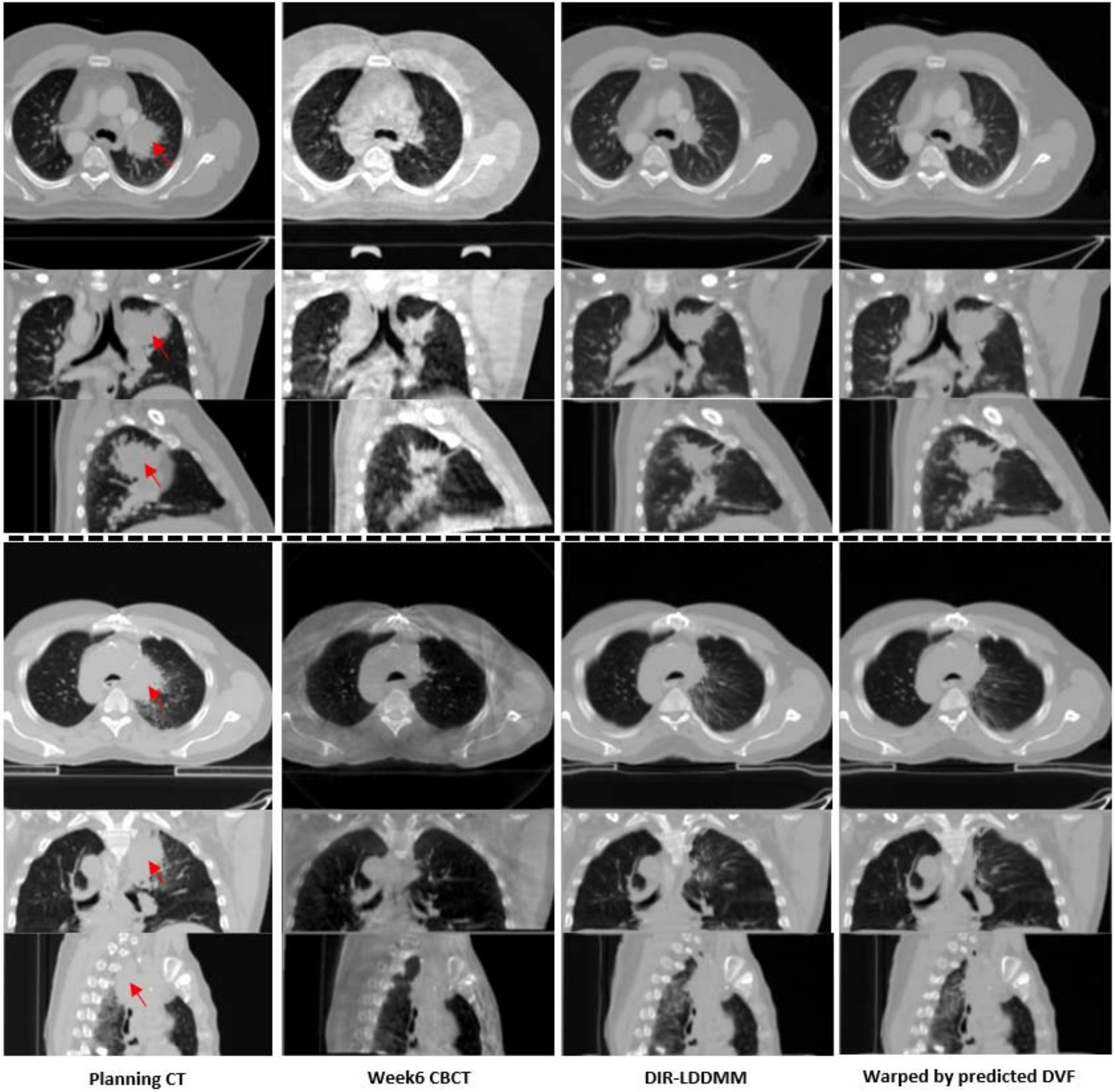}
\caption{The comparison among the planning CT, week6 CBCT, warped image by actual DVF (planning CT → week6 CBCT), and warped image by predicted DVF. The red arrow indicated the tumor region.}
\label{fig:figure9}
\end{center}
\end{figure}

{To summarize, our proposed study can effectively predict future DVFs hence it can be used for various purposes including image, contour transformation, and dose accumulation. Figure \ref{fig:figure8} is two examples of DVF and week 6 GTV contour prediction. The predicted DVF has high similarity with the actual DVF and the predicted GTV agreed well to the ground truth GTV. In addition, as shown in Figure \ref{fig:figure9}, the proposed method can predict high-quality weekly images by warping the high-quality planning CT using the predicted DVF. The GTV regions of the predicted images closely resembled the GTV region of the warped image by using the actual DVF from the registration of weekly CBCT with the planning CT.}  

There are several future directions for improving our model. First, we plan to extend ConvLSTM block to 3D. Our model used 3-channel DVFs of multiple slices as input; each slice contains 3-axis DVF. The multiple slice 3D DVFs are combined into the mini-batch structure and fed to the trained model. By replacing the 2D convolution operation of the ConvLSTM block with 3D convolution, we plan to develop a model that can train 3D DVF of the entire volume at once. {To the best of our knowledge, there are no previous studies with a complete 3D seq2seq model due to the memory constraint. We are currently extending our model to 3D using high-performance computing cluster.} Through this, we expect to predict the entire volume 3D DVF sequence instead of predicting 3D DVF for each slice. Second, we plan to apply a spatial transformation layer to apply unsupervised Seq2Seq learning. Recently, several unsupervised learning models have been proposed, such as VoxelMorph \cite{balakrishnan2019voxelmorph} for medical imaging DIR. Since unsupervised learning has the advantage of predicting DVFs from image pairs, it is expected to be able to predict DVF at each timepoint from the sequence image. Lastly, it is expected that the performance can be further improved by applying attention to features that have greater influence on the future sequence using an attention block. {In the future, we will also evaluate incorporating more recent ConvLSTM \cite{lange2020attention,su2020convolutional} and transformer-based modules (handles missing/noisy observations more naturally) \cite{giuliari2020transformer,liu2020convtransformer} in our proposed seq2seq model along with evaluating these on a much larger 1000 patient lung radiotherapy cohort we are assembling.}   

\section{Conclusion}
In this work, we presented a novel Seq2Seq model for medical images, leveraging the complete 3D imaging information of a relatively large longitudinal clinical dataset, to carry out longitudinal GTV/OAR time-series predictions for HN and lung radiotherapy patients. We also quantitatively confirmed that we can derive better longitudinal predictions using DVF rather than image representation. To the best of our knowledge, ours is the first DVF-based time-series prediction study applicable to the field of RT. The presented approach has the potential to improve RT patient outcomes by predicting the longitudinal GTV/OAR changes in advance. 

\section*{Acknowledgments}
This project was supported by MSK Cancer Center Support Grant/Core Grant (P30 CA008748).

\section*{Financial Disclosures}
The authors have no conflicts to disclose.

\end{document}